\documentclass{isprs}
\usepackage{subfigure}
\usepackage{setspace}
\usepackage{geometry}
\usepackage{epstopdf}
\usepackage[labelsep=period]{caption}
\usepackage[british]{babel}
\usepackage[hang]{footmisc}
\usepackage{booktabs}
\usepackage{multirow}
\usepackage{amsmath}
\usepackage{url}
\usepackage{graphicx}
\usepackage{xcolor}

\begin{document}

\title{UrbanVGGT: Scalable Sidewalk Width Estimation from Street View Images}
\date{}

\author{Kaizhen Tan\textsuperscript{1}, Fan Zhang\textsuperscript{2}}

\address{
  \textsuperscript{1}Heinz College of Information Systems and Public Policy, Carnegie Mellon University, USA, kaizhent@cmu.edu\\
  \textsuperscript{2}Institute of Remote Sensing and Geographical Information System, Peking University, China, fanzhanggis@pku.edu.cn
}

\abstract{
Sidewalk width is an important indicator of pedestrian accessibility, comfort, and network quality, yet large-scale width data remain scarce in most cities. Existing approaches typically rely on costly field surveys, high-resolution overhead imagery, or simplified geometric assumptions that limit scalability or introduce systematic error. To address this gap, we present UrbanVGGT, a measurement pipeline for estimating metric sidewalk width from a single street-view image. The method combines semantic segmentation, feed-forward 3D reconstruction, adaptive ground-plane fitting, camera-height-based scale calibration, and directional width measurement on the recovered plane. On a ground-truth benchmark from Washington, D.C., UrbanVGGT achieves a mean absolute error of 0.252\,m, with 95.5\% of estimates within 0.50\,m of the reference width. Ablation experiments show that metric scale calibration is the most critical component, and controlled comparisons with alternative geometry backbones support the effectiveness of the overall design. As a feasibility demonstration, we further apply the pipeline to three cities and generate SV-SideWidth, a prototype sidewalk-width dataset covering 527 OpenStreetMap street segments. The results indicate that street-view imagery can support scalable generation of candidate sidewalk-width attributes, while broader cross-city validation and local ground-truth auditing remain necessary before deployment as authoritative planning data.
}

\keywords{Sidewalk width, Street view images, 3D point cloud, Semantic segmentation, Ground-plane fitting, Metric calibration.}

\maketitle

\sloppy

\section{Introduction}\label{sec:introduction}

Sidewalk width is a fundamental micro-scale attribute of the pedestrian environment that affects comfort, safety, and accessibility. Adequate sidewalk width is especially important for wheelchair users, people with strollers, and visually impaired pedestrians, and are reflected in accessibility design standards; for example, the ADA Standards for Accessible Design specify minimum clear widths for accessible routes~\cite{adaag2010}. Despite its importance, sidewalk-width data remain scarce in most cities, limiting evidence-based pedestrian planning and infrastructure assessment.

Early efforts to build sidewalk inventories relied on field surveys or manual interpretation of imagery~\cite{proulx2015database}. Although such approaches can produce useful datasets, they are labour-intensive and difficult to scale. More recently, computer-vision methods have enabled automatic sidewalk extraction from aerial and satellite imagery. For example, TILE2NET~\cite{hosseini2023mapping} generates sidewalk polygons from overhead imagery to support large-scale sidewalk mapping. However, these approaches depend on high-resolution orthorectified imagery, which is not uniformly available across cities, and their performance may degrade under occlusion from vegetation, canopies, or parked vehicles.

Street-view imagery provides complementary ground-level detail that is often missed from overhead data~\cite{biljecki2021street}. Image collections from platforms such as Google Street View therefore offer new opportunities for sidewalk analysis. Existing street-view-based approaches, however, still face limitations. Some methods leverage provider-supplied depth information to derive measurable land-cover maps from panoramic imagery~\cite{ning2022converting}. Others estimate sidewalk width from paired street-view images using trigonometric relationships~\cite{lieu2025novel}. These methods either depend on external depth products or require multiple images together with simplified geometric assumptions that may introduce systematic error~\cite{perez2025streetscape}.

\begin{figure}[t!]
\begin{center}
\includegraphics[width=0.95\columnwidth]{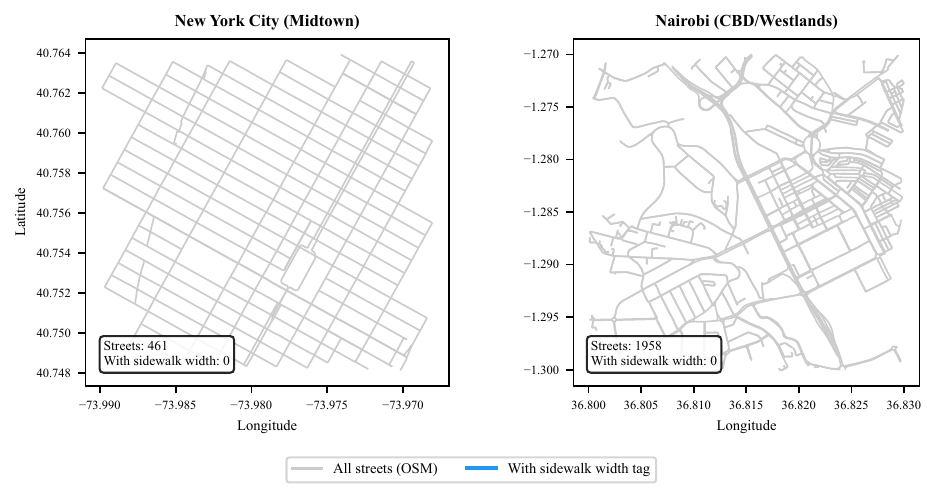}
\caption{Sidewalk width data in OpenStreetMap. Grey lines denote all drivable streets; blue lines (if any) denote streets with a sidewalk-width tag. Both New York City (461 streets) and Nairobi (1958 streets) have zero sidewalk-width tags, highlighting the data gap that UrbanVGGT aims to fill.}
\label{fig:osm_gap}
\end{center}
\end{figure}

At the same time, OpenStreetMap~\cite{haklay2008openstreetmap} provides open and globally consistent street-network data, yet sidewalk-width attributes remain largely absent. As Figure~\ref{fig:osm_gap} illustrates, none of the 461 drivable street segments in Midtown Manhattan nor any of the 1958 segments in Nairobi's central business district carries a sidewalk-width attribute in the extracted OpenStreetMap data. This gap motivates the need for a scalable way to generate candidate sidewalk-width information from widely available imagery.

To address this problem, we propose UrbanVGGT, a single-image pipeline for estimating metric sidewalk width from street-view imagery. The method combines semantic segmentation, feed-forward 3D reconstruction, ground-plane fitting, camera-height-based scale calibration, and directional width measurement on the recovered plane. By formulating sidewalk-width estimation as a plane-constrained 3D measurement problem, UrbanVGGT estimates width in reconstructed scene space rather than relying only on simplified image-plane geometry. The contributions of this paper are threefold:
\begin{enumerate}
  \item We formulate sidewalk-width estimation from street-view imagery as a plane-constrained 3D measurement problem and evaluate this formulation on a ground-truth benchmark from Washington, D.C.
  \item We compare alternative depth and reconstruction backbones within a unified downstream pipeline to isolate the effect of the geometry representation on sidewalk-width measurement accuracy.
  \item We demonstrate the feasibility of neighbourhood-scale deployment in New York City, S\~ao Paulo, and Nairobi, producing SV-SideWidth, a prototype dataset of candidate segment-level sidewalk-width estimates derived from street-view imagery.
\end{enumerate}

\section{Related Work}\label{sec:related}

\subsection{Sidewalk Measurement from Remote Sensing}

Overhead imagery has been widely used to map sidewalk and pedestrian infrastructure. Prior work developed an active transportation database through field and imagery-based collection of pedestrian-infrastructure attributes~\cite{proulx2015database}. More recently, TILE2NET was proposed as a scalable semantic segmentation approach that generates sidewalk network datasets from aerial imagery across entire municipalities~\cite{hosseini2023mapping}. These datasets can in turn support the derivation of geometric attributes such as sidewalk width. While effective in well-mapped cities with high-resolution ortho-imagery, these methods face two inherent limitations: trees and building overhangs occlude sidewalks from a nadir perspective, and suitable high-resolution imagery is often unavailable in lower-data settings.

\subsection{Sidewalk Measurement from Street-View Imagery}

Street-view imagery provides an oblique, ground-level perspective that complements overhead views. Prior work has used depth maps associated with Google Street View panoramas to convert images into land-cover maps with metric coordinates, enabling sidewalk width extraction~\cite{ning2022converting}. Another recent study estimates sidewalk width from a pair of street-view images captured at different pitch angles, applying trigonometric functions under a flat-ground assumption~\cite{lieu2025novel}. However, this approach requires two images per measurement and is sensitive to camera field-of-view settings. Vision and language models have also been explored for streetscape assessment, although reported metric accuracy remains limited~\cite{perez2025streetscape}. The broader idea of recovering metric quantities from a single image dates back to single-view metrology~\cite{criminisi2000single}. Our work builds on this line of research by replacing simplified geometric assumptions with learned 3D reconstruction and camera-height-based metric calibration, achieving higher accuracy from a single image while avoiding the need for manual vanishing-point annotation.

\subsection{Monocular Depth and 3D Reconstruction}

Monocular depth estimation has advanced rapidly. Metric-depth models such as ZoeDepth~\cite{bhat2023zoedepth}, DepthPro~\cite{bochkovskiy2024depthpro}, UniDepthV2~\cite{piccinelli2025unidepthv2}, and Metric3Dv2~\cite{hu2024metric3d} estimate metric depth from a single image. By contrast, the base versions of models such as DPT~\cite{ranftl2021vision} and Depth Anything~\cite{yang2024depth} are more commonly used for relative or affine-invariant depth prediction. Feed-forward visual geometry methods such as DUSt3R~\cite{wang2024dust3r}, CUT3R~\cite{wang2025cut3r}, VGGT~\cite{wang2025vggt}, $\pi^3$~\cite{wang2025pi3}, and MapAnything~\cite{keetha2026mapanything} infer scene geometry directly from one or more images without iterative optimisation. Among them, MapAnything is a unified metric 3D reconstruction model that can operate under multiple geometric input settings, while CUT3R is more specifically designed for continuous or sequential 3D reconstruction with persistent state. Although some of these methods are primarily developed for pairwise, sequential, or multi-view settings, several also provide a direct single-image evaluation path. In our benchmark, we exclude DUSt3R because it is most commonly used in pairwise or multi-view configurations, but retain the others for direct single-image evaluation.

These model families correspond to the three evaluation categories used in Section~\ref{sec:benchmark}: metric depth used at native scale (Category~1), monocular depth converted to 3D by pinhole unprojection and camera-height calibration (Category~2), and direct single-image point-cloud reconstruction (Category~3). Because some backbones support multiple output modalities, a small number of models, such as UniDepthV2 and MapAnything, appear in more than one benchmark category. In this work, we benchmark these model families for sidewalk-width measurement and adopt VGGT as our primary backbone because it supports single-image inference, demonstrated the most stable performance in our internal pilot experiments, and offered relatively efficient inference compared with the other candidates.

\section{Methodology}\label{sec:method}

\begin{figure*}[ht!]
\begin{center}
\includegraphics[width=0.95\textwidth]{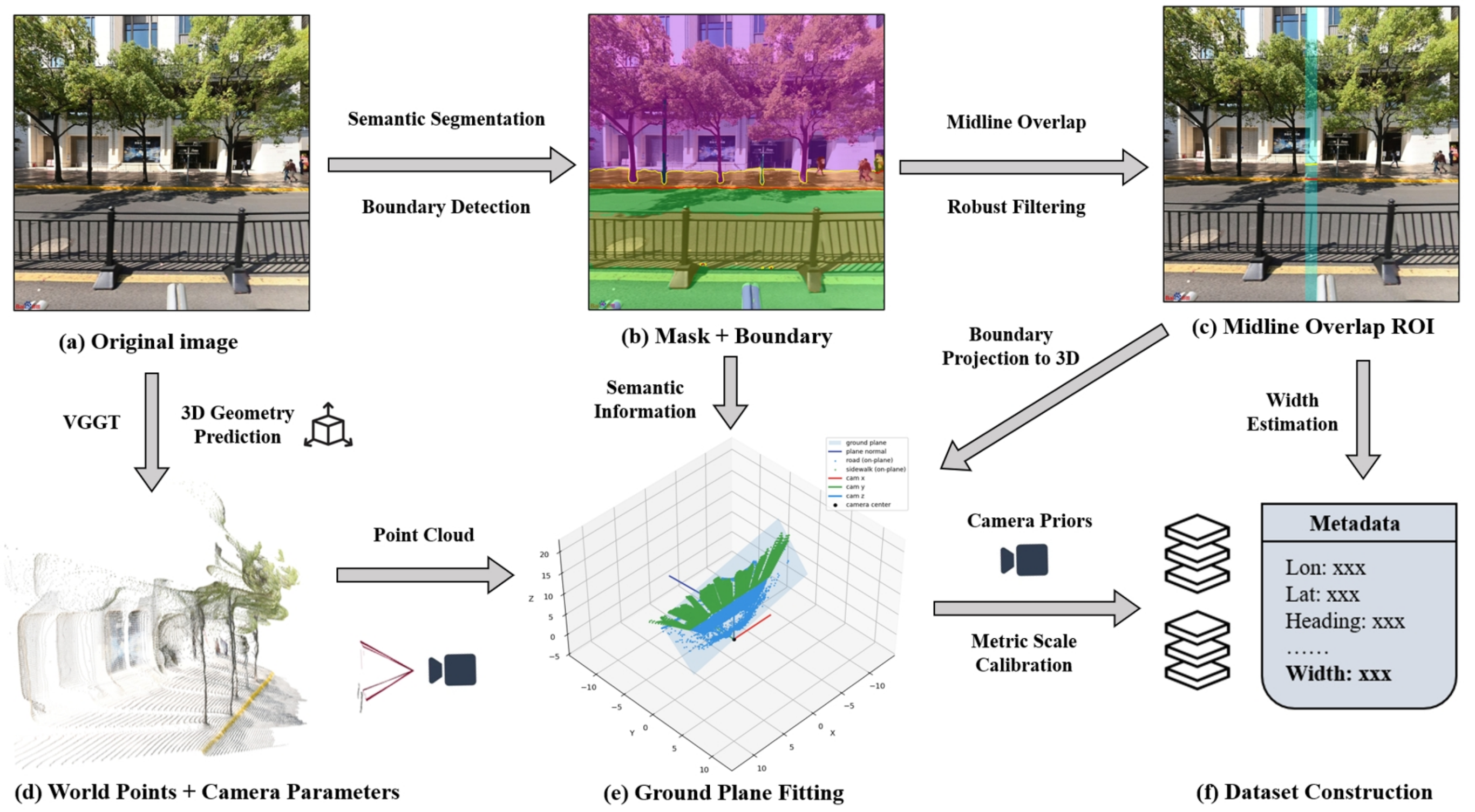}
\caption{UrbanVGGT pipeline overview. (a)~Input street-view image. (b)~Semantic segmentation with inner (yellow) and outer (red) boundary detection. (c)~Midline overlap region used to pair boundary points. (d)~VGGT-based 3D reconstruction. (e)~Ground-plane fitting with semantic point cloud. (f)~Width estimation and preliminary dataset construction.}
\label{fig:pipeline}
\end{center}
\end{figure*}

Our framework integrates five stages into a unified pipeline: semantic segmentation, 3D geometry reconstruction, ground-plane fitting, metric calibration, and robust width estimation (Figure~\ref{fig:pipeline}). Rather than treating sidewalk width as a 2D pixel distance or approximating it with image-plane trigonometry, the pipeline measures width as a directional 3D quantity on a semantically recovered ground plane, requiring only one external scalar parameter: the camera mounting height. The following subsections detail each stage.

\subsection{Semantic Segmentation}

We employ SegFormer-B5~\cite{xie2021segformer}, fine-tuned on Cityscapes~\cite{cordts2016cityscapes}, to obtain pixel-level semantic labels for urban street scenes. Among the predicted classes, we retain the \emph{sidewalk} and \emph{road} categories for downstream processing. The resulting segmentation map serves two roles in our pipeline: it identifies sidewalk boundary candidates for width estimation and provides ground-support regions for subsequent plane fitting. To improve robustness, we apply standard mask post-processing to remove small spurious regions and fill minor holes. We further discard images with insufficient sidewalk or road support, as such cases tend to produce unstable geometric estimates.

\subsection{3D Geometry Reconstruction via VGGT}

We reconstruct scene geometry using the Visual Geometry Grounded Transformer (VGGT)~\cite{wang2025vggt}, a feed-forward model that predicts dense 3D scene structure and camera parameters directly from one or more images in a single forward pass. In our single-image setting, VGGT outputs a dense per-pixel 3D point map together with the corresponding camera pose. Because these predictions are expressed in an arbitrary coordinate system and are not metric by default, a separate scale calibration step is required before physical width can be measured in metres.

\subsection{Ground-Plane Fitting}\label{sec:planefit}

We collect 3D points whose corresponding pixels are labelled as either \emph{sidewalk} or \emph{road}, and fit a ground plane
\begin{equation}
\pi: \mathbf{n} \cdot \mathbf{x} + d = 0
\label{eq:plane}
\end{equation}
where $\mathbf{n}$ is a unit normal vector and $d$ is the plane offset. To improve robustness under varying reconstruction noise, we first compute a coarse plane estimate by singular value decomposition (SVD), then estimate the dispersion of point-to-plane distances using the median absolute deviation (MAD). The inlier threshold for random sample consensus (RANSAC)~\cite{fischler1981random} is set adaptively as
\begin{equation}\label{eq:threshold}
  \tau = \mathrm{clip}\!\left(2.5 \times 1.4826 \times \mathrm{MAD},\; 0.005,\; 0.05\right).
\end{equation}
After RANSAC, all inlier points are refitted by SVD to obtain the final plane parameters $(\mathbf{n}, d)$. This adaptive strategy avoids manually tuning a single threshold and improves stability across scenes with different surface roughness and reconstruction quality.

\subsection{Metric Scale Calibration}

Because VGGT predicts geometry only up to an unknown global scale, we recover metric scale from the known camera mounting height $h_{\mathrm{cam}}$. Let $\mathbf{c}$ denote the predicted camera centre. The signed distance from $\mathbf{c}$ to the fitted ground plane gives the predicted camera height
\begin{equation}
h_{\mathrm{pred}} = |\mathbf{n} \cdot \mathbf{c} + d|
\label{eq:hpred}
\end{equation}
The metric scale factor is then computed as
\begin{equation}\label{eq:scale}
  s = h_{\mathrm{cam}} \,/\, h_{\mathrm{pred}}.
\end{equation}
All subsequent 3D measurements are multiplied by $s$ to convert VGGT coordinates into metres. In our experiments, we set $h_{\mathrm{cam}} = 2.5$\,m as a fixed prior for Google Street View car imagery, following~\cite{lieu2025novel}. Section~\ref{sec:camheight} evaluates the sensitivity of the final width estimates to this assumption.

\subsection{Column-Wise Width Estimation}

We estimate sidewalk width by scanning the sidewalk mask column-wise within a central measurement band of the image, rather than relying on morphological boundary extraction, which is often unstable in the presence of irregular masks. For each valid column, we identify the most plausible sidewalk segment and take its inner and outer boundaries as candidate sidewalk edges. These pixel locations are then mapped to the VGGT 3D point map and projected onto the fitted ground plane.

To measure width consistently across columns, we estimate the across-sidewalk direction from the projected boundary geometry and compute a per-column width as the projected distance between the two boundary points along this direction. The final sidewalk width is obtained by aggregating valid per-column measurements with robust statistics. We additionally apply a series of plausibility and consistency checks to reject unstable estimates caused by segmentation failures or geometric reconstruction errors.

\section{Experiments}\label{sec:experiments}

We evaluate UrbanVGGT on a sidewalk-width benchmark derived from Washington, D.C.\ street-view imagery. Unless otherwise stated, all methods use the same downstream pipeline, including semantic segmentation, column-wise boundary extraction, adaptive ground-plane fitting, and robust aggregation. Performance is reported using mean absolute error (MAE), root mean square error (RMSE), bias, and the proportions of estimates within 0.25\,m and 0.50\,m of the ground truth.

\subsection{Dataset and Evaluation Protocol}\label{sec:studyarea}

We evaluate the proposed method on a ground-truth sidewalk-width dataset from Washington, D.C.\ derived from the study in~\cite{lieu2025novel}. From this dataset we sampled approximately 300 Google Street View images with associated reference width measurements. All images are $640 \times 640$ pixels with a $90^{\circ}$ field of view, and the reference sidewalk widths range from 0.56\,m to 3.94\,m.

For backbone comparison, all methods share the same downstream processing and differ only in the source of 3D geometry.

\subsection{Qualitative Results}\label{sec:qualitative}

\begin{figure}[ht!]
\begin{center}
\includegraphics[width=0.95\columnwidth]{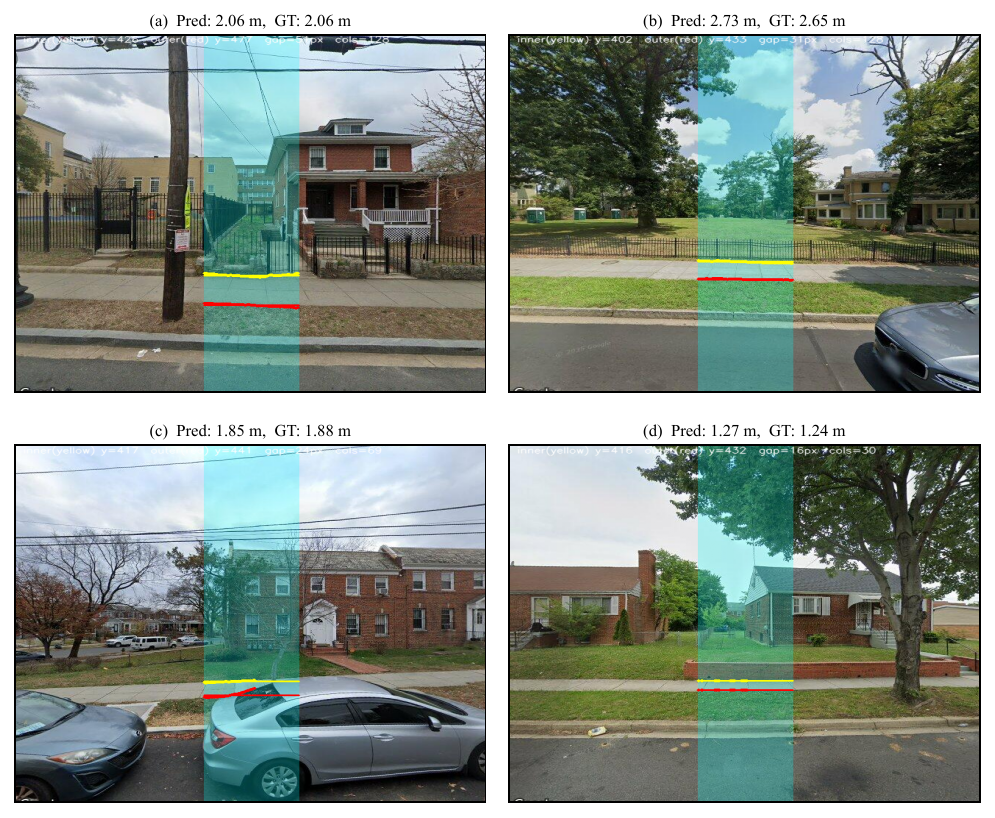}
\caption{Qualitative measurement examples on the D.C.\ dataset. Each panel shows the segmentation overlay with detected inner (yellow) and outer (red) boundaries. Predicted width: model estimate; ground-truth width: reference measurement.}
\label{fig:qualitative}
\end{center}
\end{figure}

Figure~\ref{fig:qualitative} shows representative examples from the Washington, D.C.\ benchmark. In most cases, the proposed pipeline identifies the inner and outer sidewalk boundaries reliably and produces width estimates that are visually consistent with the ground-truth measurements.

\subsection{Ablation Study}\label{sec:ablation}

We assess the contribution of the main pipeline components by selectively removing or replacing them. Table~\ref{tab:ablation} summarises the results. The full pipeline achieves the best overall accuracy, with a MAE of 0.252\,m and 95.5\% of estimates within 0.50\,m of the ground truth. Removing camera-height scale calibration causes catastrophic failure, confirming that VGGT geometry is not metrically scaled by default. Replacing the 3D reconstruction stage with pinhole geometry alone also substantially degrades performance. Finally, using the full image width instead of the central measurement band slightly worsens accuracy, suggesting that peripheral regions introduce additional noise from perspective effects and segmentation instability.

\begin{table}[ht!]
\centering
\caption{Ablation study on the Washington, D.C.\ dataset. Best values are shown in bold.}
\label{tab:ablation}
\resizebox{\columnwidth}{!}{%
\begin{tabular}{l c c c c}
\toprule
Variant & \shortstack{MAE\\(m)} & \shortstack{RMSE\\(m)} & \shortstack{$<$0.25 m\\(\%)} & \shortstack{$<$0.50 m\\(\%)} \\
\midrule
Full pipeline & \textbf{0.252} & \textbf{0.293} & 49.9 & \textbf{95.5} \\
$-$ Scale calibration & 1.571 & 1.599 & 0.0 & 0.0 \\
Pinhole only & 1.096 & 1.203 & 4.5 & 11.0 \\
Full image width & 0.265 & 0.351 & \textbf{54.2} & 88.5 \\
\bottomrule
\end{tabular}
}
\end{table}

\subsection{Camera Height Sensitivity}\label{sec:camheight}

\begin{figure}[ht!]
\begin{center}
\includegraphics[width=0.95\columnwidth]{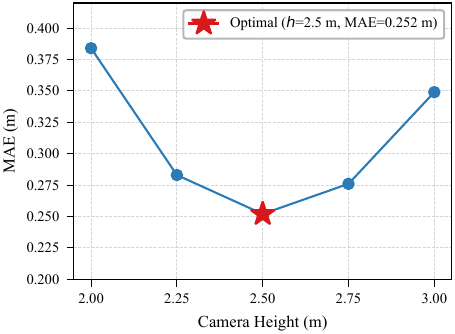}
\caption{Camera height sensitivity: MAE as a function of the assumed camera mounting height $h_{\mathrm{cam}}$.}
\label{fig:camheight}
\end{center}
\end{figure}

The camera mounting height $h_{\mathrm{cam}}$ is the only external parameter in the proposed metric calibration step. We fix $h_{\mathrm{cam}} = 2.5$\,m for Google Street View imagery throughout the study and treat it as a fixed prior rather than a tuned parameter. To assess sensitivity, we vary $h_{\mathrm{cam}}$ from 2.0\,m to 3.0\,m in increments of 0.25\,m. The results are reported in Table~\ref{tab:camheight} and visualised in Figure~\ref{fig:camheight}.

The adopted value of 2.5\,m yields the lowest MAE and the highest within-0.50\,m accuracy. Performance degrades as the assumed height departs from this value, with the sign of the bias changing accordingly. This trend is consistent with the linear scale relationship in Section~\ref{sec:method}: overestimating camera height leads to overestimation of sidewalk width, and underestimating camera height leads to underestimation. These results support the use of a fixed camera-height prior for Google Street View imagery while also highlighting the importance of setting this prior appropriately for other image providers.

\begin{table}[ht!]
\centering
\caption{Camera height sensitivity analysis.}
\label{tab:camheight}
\resizebox{\columnwidth}{!}{%
\begin{tabular}{c c c c c}
\toprule
\shortstack{Camera\\height (m)} & \shortstack{MAE\\(m)} & \shortstack{Bias\\(m)} & \shortstack{$<$0.25 m\\(\%)} & \shortstack{$<$0.50 m\\(\%)} \\
\midrule
2.00 & 0.384 & $-$0.380 & 29.7 & 65.5 \\
2.25 & 0.283 & $-$0.217 & \textbf{50.6} & 82.4 \\
2.50 & \textbf{0.252} & \textbf{$-$0.055} & 49.9 & \textbf{95.5} \\
2.75 & 0.276 & $+$0.104 & 50.0 & 85.0 \\
3.00 & 0.349 & $+$0.263 & 46.5 & 69.1 \\
\bottomrule
\end{tabular}
}
\end{table}

\subsection{Comparison with Alternative Geometry Backbones}\label{sec:benchmark}

\begin{figure*}[ht!]
\begin{center}
\includegraphics[width=0.95\textwidth]{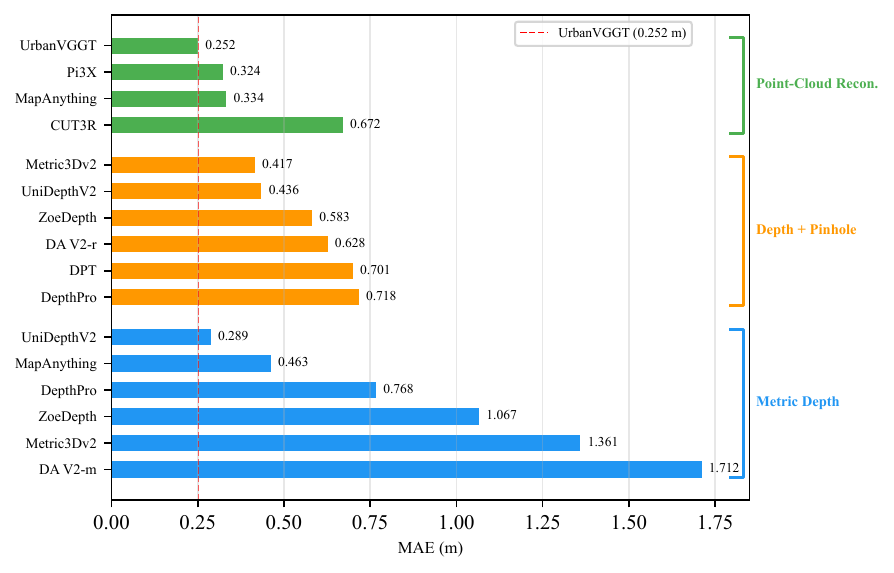}
\caption{MAE comparison across all methods. Models are grouped by evaluation category: Category~1 (metric depth with native scale), Category~2 (monocular depth with pinhole unprojection and scale calibration), and Category~3 (single-image point-cloud reconstruction with scale calibration). All methods share the same segmentation, boundary extraction, plane fitting, and outlier filtering; only the 3D geometry backbone differs. The dashed red line indicates the UrbanVGGT MAE (0.252\,m).}
\label{fig:benchmark}
\end{center}
\end{figure*}

We compare UrbanVGGT against 9 alternative depth and reconstruction models, grouped into three categories according to their geometric output representation. To ensure a fair comparison, all methods share the same downstream measurement pipeline; only the geometry backbone differs. Specifically, all models use the same SegFormer-B5 segmentation, the same column-wise boundary extraction strategy, the same adaptive plane-fitting procedure in Equation~\ref{eq:threshold}, and the same robust aggregation and rejection rules. Figure~\ref{fig:benchmark} summarises the MAE of all methods.

\subsubsection{Category~1: Metric Depth with Native Scale}

These models directly predict metrically scaled depth and are evaluated using their native geometric output, without camera-height calibration. Table~\ref{tab:cat1} shows that UniDepthV2 performs best in this category, achieving a MAE of 0.289\,m. The wide spread across models indicates that native metric depth alone does not guarantee accurate sidewalk-width estimation.

\begin{table}[ht!]
\centering
\caption{Category~1: metric depth models evaluated with native scale geometry. No camera-height calibration is applied.}
\label{tab:cat1}
\resizebox{\columnwidth}{!}{%
\begin{tabular}{l c c c}
\toprule
Model & \shortstack{MAE\\(m)} & \shortstack{$<$0.25 m\\(\%)} & \shortstack{$<$0.50 m\\(\%)} \\
\midrule
DepthAnythingV2-m & 1.712 & 0.0 & 2.7 \\
ZoeDepth & 1.067 & 10.1 & 20.6 \\
DepthPro & 0.768 & 10.2 & 23.6 \\
Metric3DV2 & 1.361 & 5.5 & 11.5 \\
UniDepthV2 & \textbf{0.289} & \textbf{54.7} & \textbf{83.3} \\
MapAnything & 0.463 & 17.8 & 60.5 \\
\bottomrule
\end{tabular}
}
\end{table}

\subsubsection{Category~2: Monocular Depth with Pinhole Unprojection}

These models predict per-pixel depth, which we unproject into 3D using camera intrinsics before applying ground-plane fitting and camera-height calibration. As shown in Table~\ref{tab:cat2}, Metric3DV2 achieves the best performance in this category, followed by UniDepthV2. Compared with direct use of native depth, this protocol improves the performance of several monocular depth models, highlighting the benefit of explicit geometric calibration.

\begin{table}[ht!]
\centering
\caption{Category~2: monocular depth models evaluated with pinhole unprojection and camera-height scale calibration.}
\label{tab:cat2}
\resizebox{\columnwidth}{!}{%
\begin{tabular}{l c c c}
\toprule
Model & \shortstack{MAE\\(m)} & \shortstack{$<$0.25 m\\(\%)} & \shortstack{$<$0.50 m\\(\%)} \\
\midrule
DepthAnythingV2-r & 0.628 & 18.0 & 38.0 \\
DepthPro & 0.718 & 12.2 & 31.7 \\
DPT & 0.701 & 15.2 & 33.7 \\
Metric3DV2 & \textbf{0.417} & \textbf{39.9} & \textbf{67.8} \\
UniDepthV2 & 0.436 & 23.9 & 66.7 \\
ZoeDepth & 0.583 & 21.0 & 44.0 \\
\bottomrule
\end{tabular}
}
\end{table}

\subsubsection{Category~3: Single-Image Point-Cloud Reconstruction}

These methods directly reconstruct dense 3D geometry from a single image. We use the reconstructed point cloud to fit the ground plane, estimate scale from camera height, and compute sidewalk width using the same measurement procedure as above. Table~\ref{tab:cat3} shows that UrbanVGGT achieves the lowest MAE in this category and the best overall performance across all compared methods. Among the alternative point-cloud methods, $\pi^3$ is the strongest competitor.

\begin{table}[ht!]
\centering
\caption{Category~3: single-image point-cloud reconstruction models evaluated with camera-height scale calibration.}
\label{tab:cat3}
\resizebox{\columnwidth}{!}{%
\begin{tabular}{l c c c}
\toprule
Model & \shortstack{MAE\\(m)} & \shortstack{$<$0.25 m\\(\%)} & \shortstack{$<$0.50 m\\(\%)} \\
\midrule
CUT3R & 0.672 & 24.3 & 47.1 \\
MapAnything & 0.334 & 44.7 & 78.0 \\
$\pi^3$ & 0.324 & 49.2 & 79.6 \\
UrbanVGGT (ours) & \textbf{0.252} & \textbf{49.9} & \textbf{95.5} \\
\bottomrule
\end{tabular}
}
\end{table}

\paragraph{Overall interpretation.}
Across all 16 evaluation configurations, UrbanVGGT achieves the lowest MAE on the benchmark. The margin over the strongest baseline is modest, however, indicating a consistent rather than dramatic improvement. In particular, UniDepthV2 also performs strongly, especially when native metric depth is available. This suggests that both strong metric-depth models and direct single-image geometric reconstruction are viable choices for sidewalk-width estimation, while UrbanVGGT provides the best overall accuracy in our setting.

\section{SV-SideWidth Feasibility Prototype}\label{sec:svwidth}

To examine whether the proposed method can help address the sidewalk-width data gaps illustrated in Figure~\ref{fig:osm_gap}, we construct SV-SideWidth, an automatically generated prototype dataset spanning three cities with different urban forms and levels of street-view availability. This deployment is intended as a feasibility demonstration rather than a validated inventory: because no ground-truth sidewalk-width data are available for these study areas, we do not report quantitative accuracy. Instead, the objective is to assess whether street-view imagery can be converted into candidate segment-level sidewalk-width attributes at neighbourhood scale, and to identify practical bottlenecks related to image availability, measurement stability, and spatial coverage.

\subsection{Study Areas}

We select three neighbourhood-scale study areas with contrasting urban contexts (Table~\ref{tab:study_areas}): Midtown Manhattan in New York City, representing a dense street grid with abundant Google Street View coverage; the Jardins/Paulista area in S\~ao Paulo, representing a high-density district with generally wide sidewalks and strong street-view availability; and the central business district together with Westlands in Nairobi, representing a setting with more limited street-view coverage. These study areas differ in urban morphology, sidewalk conditions, and image availability, making them useful test cases for prototype deployment.

\begin{table}[ht!]
\centering
\caption{SV-SideWidth study areas.}
\label{tab:study_areas}
\resizebox{\columnwidth}{!}{%
\begin{tabular}{l l l}
\toprule
City & Neighbourhood & Characteristics \\
\midrule
New York City & Midtown & Dense grid, rich Google Street View coverage \\
S\~ao Paulo & Jardins/Paulista & Dense district with relatively wide sidewalks \\
Nairobi & Central business district/Westlands & Sparser Google Street View coverage \\
\bottomrule
\end{tabular}
}
\end{table}

\subsection{Sampling and Measurement Pipeline}

For each study area, we extract the OpenStreetMap street network using OSMnx~\cite{boeing2017osmnx} based on OpenStreetMap data~\cite{haklay2008openstreetmap}. We then sample points at approximately 30\,m intervals along each road segment. At each sampled point, we estimate the local road bearing and generate two perpendicular camera headings ($\pm 90^{\circ}$) to capture the sidewalks on both sides of the street. Nearby samples are deduplicated using a 20\,m spatial grid. Google Street View images are downloaded at $640 \times 640$ pixels with a $90^{\circ}$ field of view and processed using the UrbanVGGT pipeline. Successful image-level measurements are then aggregated to the OpenStreetMap way-segment level by taking the median width of all measurements associated with the same segment.

\subsection{Prototype Coverage and Preliminary Measurements}

Table~\ref{tab:svwidth} summarises the resulting prototype dataset. Across the three study areas, the pipeline produces 1,931 valid image-level measurements, which aggregate to 527 unique OpenStreetMap street segments. Coverage is partial and varies substantially across cities, ranging from 7.6\% of segments in Nairobi to 38.2\% in New York City. This variation reflects both differences in Google Street View availability and the selectivity of the downstream quality-control procedures.

\begin{figure*}[ht!]
\begin{center}
\includegraphics[width=0.95\textwidth]{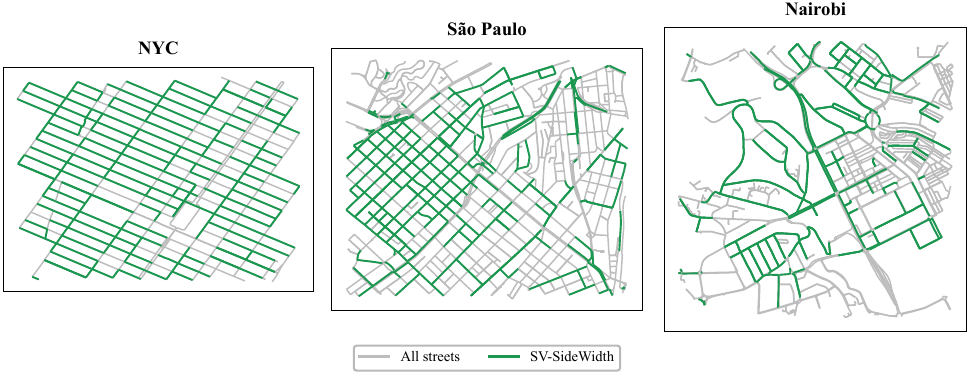}
\caption{SV-SideWidth prototype dataset coverage maps for New York City (left), S\~ao Paulo (centre), and Nairobi (right). Grey lines indicate the OpenStreetMap street network; green lines indicate street segments with SV-SideWidth measurements.}
\label{fig:coverage}
\end{center}
\end{figure*}

Within the extracted OpenStreetMap data for these study areas, none of the covered segments carries a sidewalk-width attribute. Accordingly, the prototype dataset functions as a source of new candidate width information rather than a replacement for an existing open inventory. The segment-level median widths are 2.58\,m in New York City, 2.64\,m in S\~ao Paulo, and 2.28\,m in Nairobi. These values suggest plausible cross-city variation, but they should be interpreted cautiously because the prototype has not been validated against local ground truth and does not yet achieve complete network coverage. Figure~\ref{fig:coverage} shows the spatial distribution of the measured segments over the OpenStreetMap street network. Overall, the results indicate that street-view imagery can support partial sidewalk-width data generation at neighbourhood scale, while also highlighting the current dependence on image coverage and post-measurement filtering.

\begin{table}[ht!]
\centering
\caption{SV-SideWidth prototype dataset summary and OpenStreetMap coverage.}
\label{tab:svwidth}
\resizebox{\columnwidth}{!}{%
\begin{tabular}{l c c c c c}
\toprule
City & \shortstack{Valid\\measurements} & \shortstack{Street segments\\covered} & \shortstack{Total street\\segments} & \shortstack{Coverage\\rate (\%)} & \shortstack{Median\\width (m)} \\
\midrule
New York City & 502 & 176 & 461 & 38.2 & 2.58 \\
S\~ao Paulo & 866 & 203 & 1539 & 13.2 & 2.64 \\
Nairobi & 563 & 148 & 1958 & 7.6 & 2.28 \\
\midrule
Total & 1931 & 527 & 3958 & N/A & N/A \\
\bottomrule
\end{tabular}
}
\end{table}

\section{Discussion}\label{sec:discussion}

\subsection{Complementing OpenStreetMap}

OpenStreetMap provides globally consistent topological data, including road networks and building footprints, but micro-scale pedestrian attributes such as sidewalk width remain largely absent. In the extracted OpenStreetMap data for our three study areas, none of the covered segments carries a sidewalk-width attribute (Table~\ref{tab:svwidth}). This lack of open width data limits pedestrian analysis in many cities, particularly where no authoritative sidewalk inventory exists.

Street-view imagery offers a complementary source of ground-level visual information that can be processed in a consistent manner. UrbanVGGT converts such imagery into candidate sidewalk-width estimates that can be linked to OpenStreetMap street segments without field surveys or specialised sensors. In this sense, the method is best understood as a low-cost attribute generation workflow that can support screening, prioritisation, and downstream manual review. At the same time, the current SV-SideWidth dataset remains a feasibility prototype rather than a deployable inventory: its coverage is incomplete, it has not been validated against local ground truth in the three deployment cities, and the quantitative evaluation of the method is currently limited to the Washington, D.C.\ benchmark.

\subsection{Strengths and Limitations}

The main strength of UrbanVGGT is that it estimates metric sidewalk width from a single street-view image without requiring stereo pairs, provider-supplied depth maps, or LiDAR data. The contribution is primarily a measurement pipeline that assembles semantic segmentation, feed-forward 3D reconstruction, robust plane fitting, and metric calibration into a workflow tailored to a geospatial measurement task, rather than a new architecture for visual geometry itself. This design makes the method practical for generating candidate width attributes from widely available imagery.

Several limitations, however, bound the current conclusions. First, all quantitative validation is based on a single Washington, D.C.\ dataset of approximately 300 images, which is sufficient for an initial feasibility study but not for broad claims of cross-domain robustness. Additional ground-truth datasets from cities with different street geometries, camera platforms, and sidewalk conditions are needed to assess generalisation. Second, the SV-SideWidth prototype has not been audited against ground truth in New York City, S\~ao Paulo, or Nairobi, and should therefore not be used as authoritative data for planning or policy. Third, the method depends on the spatial availability and visual quality of street-view imagery, which leads to uneven coverage across cities. Fourth, metric calibration assumes a fixed camera height for a given provider, so variation in mounting height across vehicle generations may introduce systematic bias (Section~\ref{sec:camheight}).

\subsection{Failure Modes}

Failure cases arise when the scene provides insufficient visual evidence or when local geometry violates the assumptions of the measurement formulation. One class of failure involves non-planar ground structure, such as steep driveway ramps, curb cuts, or pronounced cross-slopes, which can bias the fitted support plane. A second class involves occlusion: parked vehicles, pedestrians, street furniture, vegetation, or utility poles may obscure either the inner or outer sidewalk boundary, leading to boundary drift or unreliable measurement. A third class involves very narrow sidewalks, for which small segmentation errors can translate into relatively large width errors because only a limited number of usable columns remain within the measurement band. A fourth class involves complex curb geometry, such as bulb-outs, slip lanes, parking bays, bus stops, and shared-space streets, where the road-side and building-side boundaries are not approximately parallel and the across-sidewalk direction becomes ambiguous.

These failure modes are consistent with the behaviour observed in the benchmark and prototype experiments. In practice, the pipeline is designed to prefer conservative rejection over visually implausible measurements when the geometric evidence is weak. This behaviour is useful for large-scale candidate generation, but it also reinforces the need for downstream verification, especially in streetscape conditions that are underrepresented in the benchmark data.

\subsection{Comparison with Prior Work}

A recent method estimates sidewalk width from paired street-view images captured at two pitch angles using trigonometric relationships~\cite{lieu2025novel}. That method requires two images per measurement together with accurate camera field-of-view metadata. By contrast, UrbanVGGT operates on a single image and requires only an assumed camera mounting height for metric calibration. It also yields intermediate geometric outputs, including a dense 3D reconstruction and a fitted ground-plane model, which can support additional diagnostic checks during measurement. This simpler acquisition requirement makes the method attractive for batch processing when the objective is to generate width candidates over larger urban areas.

\subsection{Future Directions}

Future work should prioritise systematic validation of automatically generated sidewalk widths, including uncertainty calibration and targeted manual review of difficult segments. Multi-view fusion could combine measurements from overlapping images along the same street segment, potentially reducing variance and increasing spatial coverage. Temporal analysis with historical street-view imagery could support change detection in pedestrian infrastructure over time. More lightweight geometry backbones may also improve computational efficiency for large-scale deployment or field-based applications. Finally, extending the segmentation stage to distinguish sidewalk subtypes or surface conditions could enrich the generated dataset beyond width alone.

\section{Conclusion}\label{sec:conclusion}

We presented UrbanVGGT, a measurement pipeline for estimating metric sidewalk width from single street-view images. The pipeline combines semantic segmentation, feed-forward 3D reconstruction, robust ground-plane fitting, and camera-height-based scale calibration to measure sidewalk width as a directional 3D quantity on a recovered support plane. On the Washington, D.C.\ benchmark, UrbanVGGT achieves a MAE of 0.252\,m, with 95.5\% of estimates falling within 0.50\,m of the ground truth, and performs best among the compared configurations under a controlled evaluation protocol.

As a feasibility demonstration, we also generated SV-SideWidth, an unvalidated prototype sidewalk-width dataset covering 527 road segments across three cities. This prototype shows that street-view imagery can be converted into candidate segment-level sidewalk-width attributes at neighbourhood scale, while also revealing current practical constraints related to image availability, filtering, and the lack of local validation data. Overall, the study supports the promise of street-view-based sidewalk-width measurement as a low-cost complement to existing geospatial data sources, while underscoring the need for broader cross-domain evaluation and segment-level ground-truth verification before such outputs can be treated as authoritative planning data.

\section*{Acknowledgements}

Google Street View API costs for this study were self-funded by the first author. We also thank OpenStreetMap contributors for maintaining the open street-network data used in this study.

{
  \begin{spacing}{1.17}
    \normalsize
    \bibliography{references}
  \end{spacing}
}

\end{document}